\documentclass{article}
\usepackage[tikz]{KLMM/klmm}

\usepackage{multirow}
\usepackage{makecell}
\usepackage{pifont}
\usepackage{subcaption}
\usepackage{xfrac}

\title{MechMath Agent Team: LLM Driven Agents for Mathematical Research}
\author{
    Yichuan Cao\institution{State Key Laboratory of Mathematical Sciences, Academy of Mathematics and Systems Science, CAS}\institution{School of Mathematical Sciences, University of Chinese Academy of Sciences}\equalcontrib
    \and Ruichen Qiu\instref{1}\institution{School of Advanced Interdisciplinary Sciences, University of Chinese Academy of Sciences}\equalcontrib
    \and Junqi Liu\instref{1}\instref{2}
    \and Jiaqi Wang\instref{1}\instref{2}
    \and Dakai Guo\instref{1}
    \and Ruyong Feng\instref{1}\instref{2}
    \and Lihong Zhi\instref{1}\instref{2}
    \and Xiao-Shan Gao\instref{1}\instref{2}\Cauthor
}

\begin{document}

\maketitle

\begin{abstract}
AI reasoning has become a central focus in contemporary artificial intelligence, largely driven by the success of large language models.
However, mathematical research, which is characterized by non‑linear derivation paths, rigorous logical requirements, and protracted exploration cycles, poses severe challenges for existing reasoning systems.
To overcome these limitations, we present the \textit{MechMath Agent Team (MMAT)}, which is a large language model driven agent designed to serve as a co-pilot throughout the full cycle of mathematical research.
We design a tripartite Harness Architecture that decouples system responsibilities into Control, Execution, and Augmentation planes, thereby reconciling rigorous logical control with the agility demanded by open‑ended research.
Building upon this framework, we instantiate three specialized agents: a Knowledge Base Manager, a Natural Language Prover, and a Formal Language Prover, all operating in a closed loop to produce formally certified mathematical proofs.
We evaluate MMAT on open problems in Number Theory, Algebraic Complexity Theory, Differential Algebra, Operator Algebra, and Inequalities. 
Across a two-month deployment, 11 problems have been solved,  demonstrating its capacity to act as a co‑pilot throughout the entire research cycle.
The contributions are threefold: a general decoupled Harness Architecture for multi-agent mathematical reasoning, its concrete instantiation in the MMAT system, and empirical validation on a diverse suite of open problems.
\end{abstract}

\section{Introduction}

AI reasoning has been a constitutive pillar of artificial intelligence since its inception at the 1956 Dartmouth Summer Research Project~\citep{mccarthy2006proposal}. 
In recent years, it has re-emerged as a central focus, driven overwhelmingly by the proliferation of large language models (LLMs)~\citep{rossi2025aaai}. 
Through prompting strategies such as chain‑of‑thought~\citep{wei2022chain}, LLMs exhibit emergent multi‑step inference capabilities. 
Additionally, LLM‑based agents have expanded the scope of reasoning by integrating external tools and interacting with dynamic environments, supporting more deliberate and plan‑oriented problem solving~\citep{wang2024survey,huang2024understanding}. 
Together, these developments have raised new questions about the nature and evaluation of reasoning, cementing it as one of the most dynamic frontiers in contemporary AI.

Despite these advances, directly applying such reasoning capabilities to mathematical research remains particularly challenging, given its non‑linearity derivation trajectories, strict logical requirements, and protracted exploration cycles.
Conventional linear multi-agent pipelines frequently struggle with these demands due to their rigid, hard-coded workflows and vulnerability to error propagation~\citep{barrak2025traceability,qi2026individual}. 
To address these challenges, we depart from traditional frameworks and introduce the \textit{MechMath Agent Team (MMAT)}, a multi‑agent system built upon a disciplined, structurally robust Harness Architecture. 
By flexibly mounting various sub‑agents and toolchains, \textit{MMAT} coordinates multiple specialized large language models (LLMs) to work collaboratively while maintaining deterministic system states and traceable execution histories. 
In particular, it is designed to be agent‑agnostic, making it easy to integrate with current mainstream coding and reasoning agents such as Claude Code, Codex and ZCode, thereby anchoring our framework in practical, production‑grade toolchains.

Concretely, our Harness architecture adopts a tripartite paradigm that cleanly separates system responsibilities across three planes (Figure~\ref{fig:arch}), striking a balance between rigorous logical control and open-ended research agility.
First, the flexible \textit{Control Plane} operates a centralized Orchestrator grounded in two deterministic data structures: a global execution graph and a local task ledger. 
By bypassing the volatile conversational contexts of LLMs, it enforces explicit, state-driven task scheduling.
Second, the decoupled \textit{Execution Plane} eliminates context bloat and concurrency conflicts through dual physical and informational boundaries.
Sandboxed Isolated Workspaces contain agent operations, while artifact-centric File-Based Handoffs replace fragile peer-to-peer dialogue.
Finally, the \textit{Augmentation Plane} extends beyond standalone cognitive limits along two axes: human-AI co-reasoning introduces flexible interaction breakpoints for expert refinement and deadlock resolution, while stratified continual memory distills recurring reasoning errors into cross-session negative constraints, keeping the system evolvable.

Building upon this foundational harness framework, we instantiate the generic architecture as the specialized multi-agent system comprising three sovereign, domain-specialized agents (Figure~\ref{fig:overview}): the \textit{Knowledge Base Manager} (\textit{KB-Manager}), the \textit{Natural Language Prover} (\textit{NL-Prover}), and the \textit{Formal Language Prover} (\textit{FL-Prover}). 
They operate within a continuous, closed-loop paradigm:
\textit{NL-Prover} orchestrates multi-agent pipelines and tools to decompose problems and synthesize natural language proofs.
\textit{FL-Prover} acts as a logical referee by utilizing the Lean 4 compiler to mechanically verify natural language derivations.
\textit{KB-Manager} manages an object-oriented memory graph to organize and archive mathematical reasoning history.

\input{app/Tikz_overview}

We evaluate our framework’s empirical performance on several challenging mathematical domains, including Number Theory, Algebraic Complexity Theory, Differential Algebra, Operator Algebra,  and Inequality, going beyond static benchmarks to engage with open problems. 
By case studies, we demonstrate the system’s ability to carry a mathematical problem through a complete natural‑to‑formal proof pipeline and coordinate agent‑driven certification of its sub‑results.
More importantly, beyond solving individual problems, it can audit existing proofs while tracking incremental progress, engage human‑in‑the‑loop proving to establish new statements, and steer the broader project‑level exploration toward uncharted mathematical directions.
Taken together, these results confirm that \textit{MMAT} can act as a co‑pilot throughout the entire research cycle.

Over a two‑month internal experience period by authors and collaborators, \textit{MMAT} solved 11 mathematical problems (Table~\ref{tab:results} and Figure~\ref{fig:sparse_poly}), covering both longstanding open questions and entries from the 
OEIS~\footnote{The On-Line Encyclopedia of Integer Sequences. \url{https://oeis.org/}}. 
Notable examples include Katz’s question on the equivalence of generalized Airy operators~\citep{Katz1987}, the ``Challenge''~\citep{plaisted1984,SparsityChallenges} regarding the complexity of sparse polynomial divisibility tests~\citep{cao2026conphard}, the ``Challenge'' on the complexity~\citep{SparsityChallenges} of output-sensitive GCD computation~\citep{qiu2026nphard}, 
the open problem on the quasi-linearity~\citep{roche2018what} of sparse polynomial multiplication~\citep{huang2026quasilinear},
and the multivariate counterpart~\citep{HuangQL2026-ExactRoot} of the Erd\"os–Rényi conjecture asserting that the sparsity of $f^e$ entails the sparsity of $f$~\citep{Erdos1949,Schinzel1987,Zannier2008}.
Among these advancements, \textit{MMAT} either provides a rigorous proof of the key theorem or fully automates the derivation of the complete result. 
These findings underscore the capability of \textit{MMAT} in addressing frontier mathematical problems.

In summary, our contributions are threefold, and the paper follows this structure accordingly. 
First, we introduce a general harness architecture with decoupled Control, Execution, and Augmentation Planes (Section~\ref{sec:harness})
Second, We instantiate the harness framework into the MMAT multi-agent system with three specialized agents (Section~\ref{sec:mechmath}).
Finally, we evaluate its performance on open problems in Number Theory, Algebraic Complexity Theory, Differential Algebra, and Inequality (Section~\ref{sec:eval}).

\section{The Harness Architecture}
\label{sec:harness}
\input{app/Tikz_structure}
Mathematical research constitutes an inherently intricate and multifaceted endeavor. 
The non-linearity of its derivation trajectories, the stringent requirements for logical correctness, and the protracted exploration cycles collectively impose profound challenges for automated systems. 
To surmount these obstacles, we deviate from conventional linear multi-agent workflows and propose the Harness architecture, which is shown in Figure~\ref{fig:arch}. 
Serving as a disciplined and structurally robust foundation, the system orchestrates multiple specialized large language models (LLMs) through a tripartite architectural design.
This design systematically segregates responsibilities into three distinct planes:
(1) a \textit{Control Plane} for algorithmic orchestration and scheduling (blue part),
(2) an \textit{Execution Plane} for environmental isolation and containment (purple part), and
(3) an \textit{Augmentation Plane} for cognitive scaffolding and extension (green part).
This layered paradigm ensures deterministic system states and traceable execution histories while adeptly balancing rigorous logical validation, containment safety, and open-ended research agility.

\textbf{Flexible Control Plane (\S\ref{sec:control})}.
To break the rigidity of hard-coded workflows, the framework introduces a dynamic decision-making mechanism driven by a centralized Orchestrator. 
Instead of following a static path, the Orchestrator grounds its logic in two interlocking, deterministic data structures: a global \textit{Execution Graph} (DAG) for macro-level task scheduling and reactive routing, and a local \textit{Task Ledger} for strict micro-level status compliance and accountability. 
This explicit state-management process completely bypasses the volatile and implicit conversational contexts of LLMs.

\textbf{Decoupled Execution Plane (\S\ref{sec:exec})}.
To mitigate data concurrency conflicts and context bloat inherent in multi-agent collaboration, the framework establishes a decoupled execution plane governed by strict physical and informational boundaries. 
By utilizing \textit{Isolated Workspaces}, the system rigidly confines each sub-agent's write operations to local sandboxes, containing reasoning errors at their source. 
Simultaneously, through \textit{File-Based Handoffs}, the architecture abandons fragile peer-to-peer dialogues in favor of deterministic artifact transformations, drastically reducing context window pressure.

\textbf{Augmentation Plane (\S\ref{sec:aug})}.
To transcend the intrinsic cognitive ceilings and attention saturation of fully autonomous models, the framework incorporates an augmentation plane as an external scaffolding layer. 
Through \textit{Human-AI Co-Reasoning}, it establishes flexible interaction breakpoints that empower human mathematicians to qualitatively refine proofs or break logical deadlocks via structured, human-readable progress notes. 
Complementing this vertical collaboration, its \textit{Stratified Continual Memory} distills recurring high-level reasoning errors into cross-session negative constraints while externalizing dense domain facts to the underlying Knowledge Base. 
By combining targeted human intuition with lean experience retention, this plane scales the system beyond standalone automation into an evolvable research ecosystem.

\subsection{Flexible Control Plane} 
\label{sec:control}
To break the structural rigidity of hard-coded, linear pipelines, the Harness framework introduces a flexible, orchestrator-driven control flow. 
As illustrated in Figure~\ref{fig:arch}, the central \textit{Orchestrator} acts as the primary decision-making engine, dynamically determining the next system action instead of following a static, predefined path. 
To manage this execution flow without relying on the implicit, volatile conversational contexts of large language models~\citep{liu2024lost}, the \textit{Orchestrator} grounds its decision-making logic in two interlocking, deterministic data structures. 
Globally, it maintains an \textit{Execution Graph} to model macro-level task dependencies and enable reactive routing (\S\ref{sec:exec-graph}). 
Locally, it enforces a structured \textit{Task Ledger} to maintain strict micro-level status compliance and accountability (\S\ref{sec:ledger}). 
Together, these dual abstractions transform the system from a fragile, text-driven dialogue flow into an explicit, auditable state-management process.

\subsubsection{Execution Graph}
\label{sec:exec-graph}

Instead of following a rigid, linear pipeline, execution transitions are coordinated via an explicit, dependency-aware Directed Acyclic Graph (DAG) over the global task space. 
As illustrated in the Execution Graph component of Figure~\ref{fig:arch}, the \textit{Orchestrator} maps the macro-workflow topology directly onto this graph structure, where nodes represent distinct subtasks and edges delineate strict data or logical dependencies. 
Drawing on standard workflow scheduling and modern multi-agent graph paradigms~\citep{tekawade2022makespan,yu2025dyntaskmas}, this framework allows the centralized control plane to monitor execution prerequisites and structural mutations in real time, serving as the definitive baseline for global task orchestration.

Structuring the research task space as a reactive DAG yields two fundamental architectural advantages.
First, the \textit{Orchestrator} can launch parallel explorations. 
It continuously scans the topology to isolate mutually independent subtask nodes (e.g., independent lemmas of a mathematical theorem whose prerequisite data dependencies are satisfied) and seamlessly dispatch them to multiple concurrent agents, maximizing computational throughput.


Second, the DAG enables dynamic routing for error recovery.
Rather than treating reasoning missteps or validation failures as terminal exceptions, the \textit{Orchestrator} intercepts them as typed routing signals~\citep{ridnik2024code}.
Depending on the failure scope, the system adaptively alters the graph topology through three tailored strategies as follows. 
(1) For localized errors, it reroutes the faulty node to a freshly instantiated copy of the original subtask. 
(2) For broader architectural defects, it triggers a macro graph reset to upstream planning nodes to resolve structural logical inconsistencies.
(3) To bypass localized deadlocks, it spawns entirely new exploratory branches under alternative constraints.
Ultimately, this seamless integration of multi-branch parallel progression and self-healing error recovery equips the execution graph with high topological flexibility.

\subsubsection{Task Ledger}
\label{sec:ledger}
To solve tracking issues inherent in multi-round multi-agent workflows, we introduce an explicit \textit{Task Ledger} as a persistent, structured scheduling baseline that thoroughly decouples the global operational state from the implicit conversational contexts of LLMs.
At a high level, this ledger serves as a single source of truth for the system's current state, enabling the \textit{Orchestrator} to query precise status information and bypass the ``fuzzy memory'' and attention drift that plague long contexts~\citep{liu2024lost}.

Inspired by \citet{hong2024metagpt}, we design the ledger to maintain a high-fidelity record across four primitive fields (Figure~\ref{fig:arch}): task identification (\texttt{ID}), task description (\texttt{Task}), the designated responsible entity (\texttt{Owner}), and real-time execution progress (\texttt{Status}).
It is mechanically maintained by external tooling and governed by deterministic state transition rules.
To preserve state integrity and prevent unauthorized manipulation, individual subagents are denied write permissions; only the centralized Orchestrator holds the privilege to modify the ledger's fields according to strict, pre-defined structural conditions.

This centralized, constraint-driven enforcement yields several concrete architectural advantages over traditional dialogue-driven multi-agent systems.
First, by delegating state tracking to this rigid external structure, the \textit{Orchestrator} can precisely identify the current state and the next step at any moment, eliminating reliance on fallible conversational memory.
Second, the system continuously verifies field completeness and logical coherence through automated tooling, ensuring that every pending item maintains an assigned owner and every blocking state remains traceable.
Third, this constraint-based design prevents agents from using ambiguous natural language to fabricate task progress~\citep{ji2023hallucination}.
Consequently, it successfully transforms the traditionally ill-defined challenge of multi-agent accountability into an explicit, recoverable, and auditable state-management engineering process.

\subsection{Decoupled Execution Plane}
\label{sec:exec}
To complement the logical constraints enforced by the centralized control flow, the Harness establishes a decoupled execution plane governed by strict physical and informational boundaries. 
As mapped out in the \textit{Execution plane} region of Figure~\ref{fig:arch}, this architecture shifts away from traditional multi-agent paradigms that rely on shared environments and unstructured dialogue flows. 
Instead, it systematizes execution reliability through a dual-pronged approach: a spatial dimension that confines runtime file operations, and an informational dimension that structures collaborative data exchanges. 
By systematically separating where agents work from how they communicate, this execution plane effectively curbs data concurrency conflicts, context bloat, and multi-agent coordination deadlocks during long-horizon research tasks. 
To operationalize this paradigm, the system relies on two interlocking components: an \textit{Isolated Workspace} to enforce local runtime containment, and a system of \textit{File-Based Handoffs} to drive artifact-centric progression.

\subsubsection{Isolated Workspace}

Serving as a physical countermeasure against execution interference, the Harness introduces an \textit{Isolated Workspace} mechanism that establishes strict containment for each agent's file operations. 
As illustrated in the purple part of Figure~\ref{fig:arch} (bottom-left), the system allocates a dedicated, sandboxed directory to each instantiated subagent (e.g., Subagent 1, 2, and 3). 
Within this paradigm, subagents are granted strictly managed read permissions to fetch upstream context, while all write operations are rigidly confined within their local sandbox. 
Cross-zone tampering or unauthorized data modification among other subagents is strictly prevented by the execution environment.

This stringent spatial isolation yields two significant architectural advantages. 
First, it guarantees file integrity across both the parallel and sequential dimensions of the workflow. 
From a parallel perspective, the isolated workspace eliminates the risk of data concurrency conflicts, since restricting write permissions exclusively to an individual sandbox ensures that concurrent subagents never interfere. 
From a sequential perspective, this strict encapsulation prevents post-hoc alteration, unauthorized deletion, or tampering by subsequent entities. 
A downstream agent awakened in a later phase of exploration is strictly prevented by the isolation mechanism from altering or destroying the persistent historical outputs generated by its predecessors. 
This immutability is practically enforced by underlying filesystem permission controls and can be further complemented by optional content-hashing verification at the storage interface. 
Together, this multi-dimensional protection preserves a reliable, immutable audit trail where every file variant remains permanently traceable to its specific authoring entity.

Second, this isolation physically severs the pathway for unverified, faulty, or hallucinatory content to contaminate the global system state. 
In a completely open environment, a localized reasoning misstep or buggy code snippet is immediately exposed to the public workspace, where it is often repeatedly referenced, propagated and amplified by other agents~\citep{xie2026spark,barrak2025traceability}. 
By confining unverified trials within isolated directories, the Harness effectively contains reasoning faults at their source.
A localized data artifact can be delivered to peer subagents only after passing rigorous system validation and being explicitly approved by the centralized \textit{Orchestrator}.

\subsubsection{File-Based Handoffs}
To operationalize collaboration across these isolated boundaries, the Harness implements an artifact-driven workflow based on \textit{file-based handoffs}.
Subagents completely abandon fragile, dialogue-driven peer-to-peer communications and no longer pass lengthy conversational history to one another. 
Instead, as visualized by the explicit handoff in Figure~\ref{fig:arch}, subagents exchange task increments, data payloads, and acceptance criteria exclusively via structured files. 
When a subagent is awakened by the \textit{Orchestrator}, the system injects only the minimal, file-based context tailored to that specific node's local objective.
For instance, a \textit{sketcher} subagent receives a single structured file containing the initial mathematical problem and background context from the \textit{Orchestrator}, decomposes it into a collection of lemmas, and submits the report file back to the \textit{Orchestrator}. 
Crucially, the \textit{Orchestrator} does not itself inspect the content within the file. 
Instead, its responsibility is to awaken the next successor (e.g., a \textit{verifier} subagent) and transfer only the specific proposition file required for its localized task.

Decoupling agent collaboration from conversational text flows provides concrete engineering advantages:
Passing only structured files eliminates the uncontrolled growth of raw chat histories. 
This design drastically reduces the model's context window pressure, effectively mitigating the performance degradation and attention drift common in long-context language models~\citep{liu2024lost}.
More importantly, by translating vague natural-language coordination into deterministic file exchanges, the system prevents agents from getting stuck in infinite conversational loops or peer-to-peer argumentation~\citep{chu2024cohesive}. 
Since each handoff conveys a self-contained task snapshot rather than enabling bidirectional message passing, the resulting interaction graph remains acyclic and deterministic by construction. 
The entire multi-agent collaboration is thus formalized into a clean, reproducible, and losslessly preserved sequence of artifact transformations.

\subsection{Augmentation Plane}
\label{sec:aug}
While the centralized control plane governs runtime scheduling and the execution plane enforces spatial isolation, fully autonomous large language models remain fundamentally bounded by intrinsic reasoning cognitive limits and attention saturation during complex, long-horizon research tasks. 
To transcend these inherent limitations, the Harness introduces the \textit{Augmentation Plane} as an external cognitive scaffolding layer. 
As mapped out in the orange regions of Figure~\ref{fig:arch}, this plane does not interfere with the deterministic mechanics of the active control loop. 
Instead, it systematically extends the system's capabilities along two complementary axes: an interactive dimension that integrates external human intuition to refine achievements or break logical deadlocks, and a temporal dimension that distills abstract heuristic lessons across entirely separate sessions. 
By consolidating human-in-the-loop collaboration and stratified experience retention, the Augmentation Plane transforms the framework into an evolvable, highly resilient research ecosystem.

\subsubsection{Human-AI Co-Reasoning}
To overcome the inherent limitations of full autonomy~\citep{srivastava2023beyond}, the Harness implements a human–AI co-reasoning mechanism featuring flexible interaction checkpoints, which is a design mandated by two distinct operational realities. 
First, even when an agent completes a task successfully, human expert oversight remains essential.
Autonomous generation frequently requires post-processing or optimization to reach publication-ready quality, demanding qualitative refinement, alternative reasoning pathways, or structural enhancements. 
Second, when an autonomous agent encounters an insurmountable logical impasse, it lacks the intrinsic capacity to resolve the deadlock~\citep{huang2024large}.
Thus, it necessitates the intervention of a human mathematician to diagnose the underlying bottleneck and prescribe a strategic course to advance the exploration frontier.

The co-reasoning relies on two core technical mechanisms. 
First, to facilitate seamless expert comprehension, the system automatically generates a highly human-readable diagnostic report. 
Even when a task cannot be fully completed, it compiles a comprehensive \textit{progress note} that precisely logs all intermediate milestone achievements, explicit blocking coordinates, and auxiliary context. 
Second, the interface supports flexible human input modalities, permitting mathematicians to inject strategic guidance or manual corrections either by issuing direct instructions or by precisely annotating the report itself. 
The centralized \textit{Orchestrator} then ingests these critical human hints, updates the underlying task structures, and seamlessly resumes DAG exploration directly from the recovery checkpoint. 
By salvaging prior computation, this interface minimizes human effort and eliminates wasteful full-scale restarts.

\subsubsection{Stratified Continual Memory}
To sustain cross-session learning without inducing cognitive bloat~\citep{packer2024memgpt}, the Harness introduces a stratified memory architecture.
This tripartite design partitions information based on its scope of validity and cognitive density, preventing the active context from being choked by either transient noise or heavy data payloads. 
Specifically, \textit{local memory} manages intra-session dynamics by tracking temporary notation conventions (e.g., assigning a specific semantic definition to symbol $\mathcal{A}$ within a localized proof) that are vital for immediate consistency but obsolete once the session terminates. 
In contrast, \textit{long-term memory} operates cross-session to preserve abstract, highly distilled strategic boundaries that govern broader heuristic reasoning (e.g., heuristic thresholds such as differentiating between $O(T)$ and soft-$O(T)$ complexity regimes in asymptotic analysis). 
Crucially, dense and large-scale domain knowledge (e.g., certifying whether a specific concrete algorithm satisfies soft-$O(T)$ bounds) is completely externalized from the active memory apparatus and stored as persistable data artifacts within the broader Knowledge Base, ensuring that heavy factual payloads never saturate the context window.

Focusing the system's memorization capacity on capturing abstracted errors yields significantly greater architectural value than passively recording successful factual trajectories. 
While valid mathematical derivations are highly problem-specific and rarely transferable across disparate research domains, systemic reasoning failures and structural pitfalls expose universal behavioral flaws. 
By distilling these recurring high-level patterns, such as a persistent tendency to confound highly similar notation conventions inherited from entirely separate problem sessions, the memory layer transforms idiosyncratic mistakes into actionable \textit{negative constraints}, which are explicit behavioral boundaries that the system should avoid crossing in future reasoning paths~\citep{su2026mistake}. 
Prioritizing these abstract behavioral boundaries over bulky domain facts fundamentally ensures that the memory apparatus remains structurally lean, maximizing proactive defensive utility and preventing regression while minimizing context-window overhead.
Consequently, by persistently applying these negative constraints, the system aims to neutralize repetitive cognitive traps without inducing attention drift.

\section{MechMath Agent Team}
\label{sec:mechmath}
\input{app/Tikz_agentteam}

This section delineates the architectural design and operational mechanics of the \textit{MechMath Agent Team}, a highly modularized, self-correcting multi-agent system engineered to automate complex mathematical research.
We instantiate the generic Harness architecture introduced in Section 2 by specializing the template with component-specific tools and sub-agent roles. 
This process yields three sovereign, domain-specialized agents: the \textit{Natural Language Prover} (\textit{NL Prover}), the \textit{Formal Language Prover} (\textit{FL Prover}), and the \textit{Knowledge Base Manager} (\textit{KB Manager}), which will be introduced in the following sections respectively.

Rather than functioning as a rigid, linear pipeline, these agents are orchestrated within a continuous, closed-loop paradigm as depicted in Figure~\ref{fig:mechmath_workflow}. 
The architecture is instantiated as a collection of strictly isolated projects, each dedicated to a concrete research direction (e.g., sparse polynomial in Section~\ref{sec:eval_prod}). 
Within a single project, the three coequal agents share a unified object-oriented memory graph and interact organically through a tightly coupled loop of four primitives: \textit{Inform}, \textit{Formalize}, \textit{Feedback}, and \textit{Archive}. 
Different projects remain entirely disconnected to avoid data proliferation and cross-domain interference, so that each instance of this self-reinforcing interplay remains tightly focused within its designated research direction.

Concretely, \textit{Inform} flows from the \textit{KB Manager} to both the \textit{NL Prover} and the \textit{FL Prover}, proactively supplying them with relevant historical knowledge, proven lemmas, and accumulated heuristics from the shared memory graph to guide ongoing reasoning. 
The \textit{NL Prover} then navigates complex reasoning spaces, synthesizes natural language proofs, and, when a derivation requires rigorous validation, transmits it to the \textit{FL Prover} via the \textit{Formalize} primitive. 
The \textit{FL Prover} formalizes the derivation into Lean~4 code, performs deterministic mechanical verification, and returns the verification outcomes to the \textit{NL Prover} through \textit{Feedback}. 
If the verification fails, the \textit{NL Prover} revises the informal proof based on the feedback and may re-initiate the \textit{Formalize} process, thereby embodying the system’s self-correcting nature.
Throughout this process, both the \textit{NL Prover} and the \textit{FL Prover} deposit their newly generated exploratory heuristics, formalized artifacts, verification logs, and failure records into the \textit{KB Manager} via the \textit{Archive} primitive, continuously enriching the shared memory graph. 
This mutually reinforcing circulation ensures that informal insights, exploratory heuristics, verification outcomes (both successful and failed), and formalized Lean~4 artifacts are systematically consolidated, enabling accumulated mathematical experience to seamlessly seed and optimize subsequent research iterations.

\subsection{Natural Language Prover}
\label{sec:nl}
As the core agent of \textit{MMAT} responsible for generating natural language proofs, the \textit{Natural Language Prover (NL-Prover)} implements a dual-layered architecture comprising the Proof Pipeline and the Command-Line Tooling Layer to transition automated mathematical reasoning from isolated generation to a rigorous, scalable discovery process.
The \textit{Proof Pipeline} establishes a hierarchical, multi-agent framework that orchestrates specialized subagents across four functional layers—balancing baseline problem decomposition with advanced diagnostic path recovery and publication-grade documentation. 
Serving as the bedrock for this cognitive workflow, the \textit{Command-Line Tooling Layer} acts as a deterministic ``exoskeleton'' that bridges the gap between stochastic language generation and formal mathematical rigor. 
By offloading resource-intensive and mechanical tasks—such as literature retrieval, PDF parsing, and structural schema gating—to reliable programmatic routines, this infrastructure layer minimizes hallucination and preserves the agents' cognitive capacity for pure logical synthesis. 
Together, these two components form a robust synergy that allows highly divergent mathematical explorations to advance within a controllable, verifiable, and resource-efficient framework.

\subsubsection{Proof Pipeline}
Leveraging the flexible control plane of the Harness framework, the proof pipeline is structured as a hierarchical, multi-agent framework characterized by high topological flexibility and dynamic fault tolerance.
The subagents are organized into four interconnected functional layers, which are summarized in Table~\ref{tab:nlprover-role}: 
(1) the \textit{Core Derivation Baseline}, which operationalizes standard problem decomposition and verification paradigms; 
(2) the \textit{Grounded Infrastructure Interfaces}, providing empirical grounding through external database retrieval and programmatic execution; 
(3) the \textit{Advanced Mathematical Reasoning} cluster—our primary architectural innovation—engineered for non-linear path recovery, semantic auditing, and structural diagnostics; and 
(4) the \textit{Documentation} layer, which synthesizes internal reasoning traces into publication-grade artifacts. 
Orchestrated via an adaptive, state-aware control loop, this stack balances exploratory autonomy with rigorous computational and logical safeguards.

\begin{table}[t]
    \centering
    \small
    \begin{subtable}{\linewidth}
    \subcaption{Specialist Roles in NL Prover}
    \label{tab:nlprover-role}
    \begin{tabular*}{\linewidth}{@{\hspace{10pt}\extracolsep{\fill}} l l @{\hspace{10pt}}}
    \toprule
    \textbf{Role} & \textbf{Responsibility} \\
    \midrule
    Sketcher & Converts original problem into structured proof plan with lemmas and dependencies.\\
    Generator & Produces detailed informal proofs with justifications for assigned statements.\\
    Verifier & Independently audits proofs for correctness and completeness.\\
    \midrule
    Searcher & Traces named theorems to usable, verified statements for proof use.\\
    Code Executor & Programmatically validates computations, finite enumerations, and exhaustive case evidence.\\
    \midrule
    Auditor & Clarifies notations and assumptions to prevent conceptual ambiguity and logic errors.\\
    Explorer & Proposes diverse proof strategies when standard routes fail.\\
    Synthesizer & Synthesizes high-potential proposals into actionable branches.\\
    Regulator & Classifies proof failures to identify responsible owners and next steps.\\
    CE-Hunter & Investigates boundary cases and potential falsities to harden the proof search.\\
    \midrule
    Refiner & Shortens or improves verified proofs while keeping original as fallback.\\
    Writer & Compiles refined proof components into cohesive, publication-grade mathematical drafts.\\
    \bottomrule
    \end{tabular*}
    \end{subtable}

    \begin{subtable}{\linewidth}
    \subcaption{Command-Line Tools}
    \label{tab:nlprover-tool}
    \begin{tabular*}{\linewidth}{@{\hspace{10pt}\extracolsep{\fill}} l l l @{\hspace{10pt}}}
    \toprule
    \textbf{Tool} & \textbf{Interfaces} & \textbf{Description} \\ 
    \midrule
    
    Retrieval & \texttt{arxiv}, \texttt{matlas} & Retrieves verified theorems and proofs to enable cumulative reasoning. \\ 
    
    Extraction & PDF Parser & Converts unstructured visual documents into structured text segments. \\ 
    
    Linting & Linters & Enforces strict morphological and schema compliance on agent outputs. \\ 
    
    Verification & LLM API & Dispatches reasoning snippets to heterogeneous models for blind-testing. \\ 
    \bottomrule
    \end{tabular*}
    \end{subtable}
    \caption{Natural Language Prover. (a) shows four kinds, 12 different subagents in the Natural Language Prover. (b) shows 4 command-line tools and their interfaces.}
\end{table}

\textbf{Core Derivation Baseline.}
Drawing on established mathematical agent architectures, the Core Proof Pipeline integrates two foundational paradigms: 
(1) \textit{problem decomposition}, where complex tasks are broken down into actionable sub-problems, and 
(2) a \textit{generate-verify} workflow, where proof candidates are subjected to independent scrutiny~\citep{huang2025winning}. 
We realize these paradigms through the \textit{Sketcher}, which maps problems into structured blueprints and lemma taxonomies, and the \textit{Generator-Verifier} pair, which manages the production and auditing of proof attempts.

Distinct from rigid, linear implementations of these paradigms, our architecture is governed by an \textit{Orchestrator}-driven dynamic control flow. 
The \textit{Orchestrator} maintains global state visibility, allowing the system to deviate from pre-planned trajectories based on real-time feedback. 
For instance, upon a \textit{Verifier} failure, the \textit{Orchestrator} dynamically determines the optimal remediation strategy: it may instruct the \textit{Generator} to refine a specific local segment, or, in instances of fundamental logical conflicts, trigger a backtrack to the \textit{Sketcher} to adjust the underlying decomposition strategy. 
This recursive feedback loop ensures that the system proactively adapts to logical inconsistencies rather than blindly propagating errors.

\textbf{Grounded Infrastructure Interfaces.}
To anchor abstract reasoning in verified literature and computational evidence, the system employs two primary external interfaces. 
The \textit{Searcher} bridges the gap between the proof process and academic databases, ensuring that all imported theorems are contextually grounded and their preconditions formally documented. 

Concurrently, the \textit{Code Executor} serves as a programmatic fallback for tasks requiring objective validation, with a primary focus on integrating the capability to invoke symbolic computation engines. It can execute algebraic derivations, identity verifications, and polynomial simplifications directly at the level of pure symbols and exact rationals. By executing these specialized symbolic verification scripts, the \textit{Code Executor} provides an empirical "hard-science" layer, ensuring that critical proof claims rest firmly on audited, exact symbolic computational results rather than heuristic reasoning.

\textbf{Advanced Mathematical Reasoning.}
Positioned above the infrastructure layer, this cluster constitutes the core cognitive and diagnostic engine of our architecture, specifically engineered to navigate advanced mathematical nuances, execute non-linear path recovery, and perform structural failure analysis. 
As a foundational safeguard within this cognitive core, the \textit{Auditor} explicitly formalizes and clarifies specialized notations, convention-based assumptions, and localized naming protocols within the workspace. 
This rigorous semantic anchoring prevents the downstream pipeline from masking logical errors within shifting or ambiguous conceptual definitions.

When the primary pipeline encounters verification deadlocks or deep heuristic ambiguity, this cluster initiates a dynamic path-recovery sequence. 
A swarm of \textit{Explorer} agents is deployed to propose alternative mathematical strategies from multiple directions under varied macro-constraints, maintaining high search-space diversity. 
This divergent exploration is immediately distilled by the \textit{Synthesizer}, which evaluates and ranks the generated proposals. 
By converting only the most viable proposals into active speculative branches, the \textit{Synthesizer} prevents the system from persisting on failed trajectories and substantially reduces the \textit{Orchestrator}’s analytical overhead.

Complementing this exploration, the \textit{Regulator} provides deep structural diagnostics upon verification failures~\citep{an2026qed}. 
Rather than synthesizing new proof segments, it evaluates the failure state to isolate the exact root cause—whether it stems from a localized proof gap, an invalid lemma formulation, or a definition mismatch—thereby identifying the minimal responsible party for targeted remediation. 
Simultaneously, the \textit{Counterexample Hunter} (\textit{CE-Hunter}) aggressively hardens the reasoning process against unprovable propositions. 
Through logical deduction or algorithmic script execution, the \textit{CE-Hunter} actively uncovers boundary-case obstructions and falsifies flawed intermediate claims, ensuring the system promptly abandons structurally unviable paths.

\textbf{Documentation.}
The final stage of the derivation lifecycle centers on the Documentation cluster, which distills heterogeneous reasoning assets into structured, reader-facing academic artifacts. 
The \textit{Refiner} performs high-level structural compression, distilling verbose derivations into concise, logically sound arguments while utilizing a blind-verification fallback to ensure formal equivalence with the original proof. 
Subsequently, the \textit{Writer} compiles these segments into polished academic drafts. 
By managing stylistic alignment, citation formatting, and logical flow, the \textit{Writer} transforms disjointed intermediate \texttt{markdown} files into cohesive, publication-grade mathematical prose. 
This cluster ensures that the system’s accumulated internal explorations are effectively translated into transmissible knowledge, ready for human peer review or external formalization.

\subsubsection{Command-Line Tooling Layer}

Beyond the collaborative workflow of the agents, the system relies on a deterministic Command-Line Tooling layer as its underlying infrastructure, as summarized in Table~\ref{tab:nlprover-tool}. 
The core design philosophy is to bridge the gap between stochastic language generation and the absolute rigor required for mathematical proving. Rather than burdening generative models with mechanical executions, this layer acts as a "deterministic exoskeleton." 
It offloads non-reasoning tasks to reliable software routines, ensuring that the agents' cognitive capacity is strictly reserved for high-level logical synthesis.

\textbf{Literature Retrieval.}
Mathematical proving cannot occur in a vacuum, nor should computational resources be wasted on re-proving established results.
To enable cumulative reasoning, we equip the system with a dedicated retrieval pipeline that combines two complementary tools: Matlas~\citep{ju2026matlas}, a semantic search engine tailored for published mathematical documents, and the official arXiv API, which provides real-time access to preprints and their metadata.
Because mathematical definitions and theorem boundaries must be absolutely precise, relying on a model's internal parametric memory risks fatal hallucinations.
Grounding the agent's exploration in trusted external repositories ensures that any introduced lemma is historically verified, preventing the system from building proofs on fabricated mathematical premises.

\textbf{Local Reference Extraction.}
Academic literature in mathematics is uniquely dense, characterized by intricate notation, multi-column layouts, and heavy cross-referencing. 
Feeding raw, unstructured visual documents directly into an LLM causes severe token bloat and dilutes the model's attention. 
This extraction tool is critical for context window optimization. 
By converting local reference materials into an instantly queryable, structured text segment database, it allows agents to perform granular, needle-in-a-haystack lookups. 
This preserves the model’s limited context window for multi-step reasoning rather than document parsing.

\textbf{Structural Lint Gates.}
A robust math prover requires a strict separation between structural syntax and mathematical logic. 
The linting tool enforces this decoupling by acting as an automated gatekeeper. 
Before a proof draft is even evaluated for logical correctness, it must pass a battery of deterministic shape checkers. 
If an artifact violates the expected schema or communication protocol, it is rejected immediately. 
This engineering discipline prevents the system from wasting expensive model calls and reasoning cycles on syntactically malformed outputs, ensuring the entire workflow advances on a predictable track.

\textbf{Cross-model Verification.}
While the tooling layer consists of deterministic code, it serves as a reliable broker to orchestrate heterogeneous language models. 
In formal deduction, a single generative model is highly susceptible to confirmation bias—frequently overlooking its own subtle logical gaps or false deductions. 
This verification framework programmatically introduces cognitive diversity. 
By routing proof snippets through independent peer-review and blind-testing protocols, the system breaks individual reasoning traps and exposes hidden vulnerabilities, shifting the proof process from isolated generation to rigorous multi-model consensus.

\subsection{Formal Language Prover}
\label{sec:fl}
The Formal Language Prover is the core engine for achieving mathematical rigor~\citep{liu2026numina}. 
Unlike natural language systems that rely on large models for cross-verification, mathematical validity here is absolutely and exclusively determined by the Lean 4 compiler kernel in a deterministic manner. 
Anchored by this robust mechanism, the formalization prover exhibits powerful, multi-dimensional core capabilities: it can not only precisely formalize mathematically stated problems into rigorous code statements, but also autonomously explore and generate complete formal proofs under stringent compilation constraints. 
More importantly, it acts as the ultimate logical referee to assist in verifying informal proofs. 
By translating the natural language derivation drafts of large models or human experts into Lean code for mechanical verification, it completely eradicates logical illusions and loopholes hidden by the ambiguity of natural language.

\subsubsection{Formalization Architecture}

\begin{table}[t]
    \centering
    \small
    \begin{subtable}{\linewidth}
    \subcaption{Specialist Roles in FL Prover}
    \label{tab:flprover-role}
    \begin{tabular*}{\linewidth}{@{\hspace{10pt}\extracolsep{\fill}} l m{.83\linewidth} @{\hspace{10pt}}}
    \toprule
    \textbf{Role} & \textbf{Responsibility} \\
    \midrule
    Formalizer & Translates source text into Lean statement scaffolds with bodies marked as \texttt{sorry}.\\
    F-Reviewer & Validates statement and definition fidelity against the source text to ensure semantic alignment.\\
    F-Generator & Executes targeted proof derivations for specific theorems or lemmas within isolated workspaces.\\
    Integrator & Merges verified scratch proofs and artifacts into the master repository under structural constraints.\\
    Golfer & Refines post-verification code for improved aesthetics and maintainability without altering logic.\\
    Regulator & Conducts systemic global audits to detect formalization traps and manage reasoning cycle state.\\
    \bottomrule
    \end{tabular*}
    \end{subtable}

    \begin{subtable}{\linewidth}
    \subcaption{Command-Line Tools}
    \label{tab:flprover-tool}
    \begin{tabular*}{\linewidth}{@{\hspace{10pt}\extracolsep{\fill}} l m{.83\linewidth} @{\hspace{10pt}}}
    \toprule
    \textbf{Tool} & \textbf{Description} \\ 
    \midrule
    Checking & Compile and verify Lean files as the ground truth for mathematical correctness. \\ 
    
    Scanning & Scans source code for \texttt{sorry} or \texttt{admit} to ensure proof integrity. \\ 
    
    Guarding & Snapshots theorem statements to detect and prevent unauthorized AST-level tampering. \\

    Indexing & Extracts dependencies and structural outlines to facilitate proof reuse.\\
    
    Searching & Use \texttt{Leansearch}, \texttt{Leanfinder}, \texttt{Leandex} and \texttt{Loogle} to retrieve lemmas from extensive libraries. \\ 
    \bottomrule
    \end{tabular*}
    \end{subtable}
    \caption{Formal Language Prover. (a) shows six different subagents in the Formal Language Prover. (b) shows five command-line tools and their interfaces.}
\end{table}

The transition from natural-language mathematical proofs to machine-verifiable code (e.g., Lean 4) introduces severe cognitive and engineering challenges. 
Unlike informal reasoning, formalization is highly sensitive to the \textit{semantic gap}, where a slight translation error fundamentally changes a theorem's meaning. 
Additionally, it is susceptible to \textit{workspace contamination}, where exploratory proving corrupts the global file environment.
Furthermore, formal verification systems are susceptible to silent anti-patterns, such as vacuous proofs resulting from contradictory hypotheses. 
To systematically insulate the codebase from these vulnerabilities, our framework decouples the formalization lifecycle into specialized agent roles, summarized in Table~\ref{tab:flprover-role}.

To bridge the gap between informal mathematics and formal types, where language models often suffer from translation drift, the system decouples syntax generation from semantic auditing. 
The \textit{Formalizer} drafts initial Lean scaffolds, while the \textit{Formal Reviewer} (\textit{F-Reviewer}) acts as a dedicated gatekeeper. 
Requiring an explicit approval from the \textit{F-Reviewer} before a statement snapshot is established ensures that formalized theorems remain strictly aligned with the source text before any proof efforts begin.

To mitigate workspace contamination and dependency rot caused by chaotic trial-and-error exploration, the architecture enforces a strict isolation paradigm. 
The \textit{Formal Generator} (\textit{F-Generator}) conducts speculative proof searches exclusively within ephemeral scratchpads, shielding the baseline repository from corruption. 
Once a proof is validated, the \textit{Integrator} serves as the sole interface permitted to merge fragments into the target file, acting as structural glue that sanitizes namespaces without altering proof logic.

To combat code entropy and silent anti-patterns like vacuous truths, we introduce specialized optimization and auditing layers. 
The \textit{Golfer} performs conservative, post-hoc syntax cleanups to maximize human readability without compromising formal correctness. 
Concurrently, the \textit{Regulator} conducts global audits at the end of each reasoning wave, analyzing task ledgers and repository diffs to detect systemic traps like statement drift or duplicate definitions.

\subsubsection{Command-Line Tooling Layer}
In this formal environment, agents are empowered by a specialized suite of command-line tools in Table~\ref{tab:flprover-tool}. 
Since correctness is deterministically adjudicated by the Lean kernel, our tooling layer acts as a series of final-verdict gates, categorized by their primary objectives as follows.

\textbf{Verification.}
This suite of tools maintains the rigorous boundary of the formal environment. 
\textit{Checking} acts as the system's "supreme oracle," invoking the Lean compiler to provide a deterministic ground truth that eliminates model hallucinations. 
\textit{Scanning} safeguards the proof chain by auditing source code for unfinished placeholders, ensuring that no \textit{sorry} or \textit{admit} keywords bypass formal validation. 
Complementing this, \textit{Guarding} prevents derivation drift by maintaining snapshots of theorem statements and monitoring Abstract Syntax Tree (AST) integrity, which effectively blocks agents from surreptitiously weakening propositions to force a successful, yet invalid, compilation.

\textbf{Discovery.}
To handle the complexity of massive libraries, these tools optimize the agents' navigation and integration capabilities. 
\textit{Indexing} provides crucial workspace awareness by extracting dependency graphs and structural outlines, allowing agents to seamlessly reuse existing lemmas. 
Building upon this, \textit{Searching} offers a multi-dimensional retrieval ecosystem, including \texttt{leansearch}~\citep{gao2026leansearch} (mixed natural language and code search), \texttt{leanfinder}~\citep{lu2026leanfinder} and \texttt{leanexplore}~\citep{asher2025leanexplore} (semantic matching search), and \texttt{loogle} (type signature-based pattern matching)
By integrating diverse strategies, it enables agents to precisely anchor relevant premises, whether they are navigating from a target statement or a specific proof state.

\subsection{Knowledge Base Manager}
\label{sec:kb}
The Knowledge Base Manager (KB Manager) serves as the persistent memory and structural repository of the harness, enabling the system to manage complex mathematical research beyond the constraints of transient execution. 
Departing from the ``LLM-as-a-Wiki'' paradigm, which treats knowledge as unstructured text, the KB Manager re-architects the repository into an actionable, object-oriented memory graph. 
Through a hub-and-spoke architecture, it coordinates a suite of specialist agents to transform raw inputs into a rigorously verified, formalization-aware knowledge structure. 
By decoupling global orchestration from localized execution, the KB Manager ensures that the system’s reasoning history—ranging from successful proofs to verified obstructions—remains transparent, auditable, and unconditionally trustworthy.

\subsubsection{Persistent Wiki Knowledge Base}
\textit{KB Manager} adopts the ``LLM-as-a-Wiki'' paradigm~\citep{karpathy2026wiki}, which organizes a persistent knowledge base around three canonical page types: \texttt{Source} pages that capture raw input materials, \texttt{Concept} pages that distill generalized insights, and \texttt{Analysis} pages that document analytical procedures. 
Yet standard implementations of this paradigm store the content within these pages merely as unstructured, enriched notes, making them ill-suited for the precise retrieval and compositional demands of mathematical reasoning.
To meet the rigorous requirements of mathematical workflows, the \textit{KB Manager} re-architects the primitive memory unit by compiling the content of these wiki pages into a suite of strongly-typed, modular card entities.
This structural refinement decouples invariant mathematical objects (such as definitions and proven theorems) from the empirical heuristics, incremental proof fragments, and falsified constraints generated during dynamic proof search. 
As a result, the repository transforms from a passive notebook into an actionable, object-oriented memory graph that enables granular knowledge retrieval, compositional reasoning, and the systematic accumulation of exploratory experience.


Beyond optimizing the storage of natural language knowledge, the \textit{KB Manager} extends its architecture to bridge the gap between human-readable mathematics and formal verification.
The gap between natural language reasoning and machine-checkable formal representations poses a major obstacle for automated agents operating in interactive theorem proving (ITP) environments~\citep{jiang2023draft}.  
To overcome this problem, the system introduces \texttt{Lean} cards as a dedicated formalization interface.  
These cards ground mathematical concepts in concrete, machine-verifiable artifacts by explicitly tracking formal declaration names, type signatures, and proof statuses within the Lean environment.

Furthermore, mathematical discovery inherently relies on dynamic, non-linear proof-search trajectories. 
To capture this exploratory process, the \textit{KB Manager} introduces a separate cognitive memory layer comprising \texttt{PartialProof} and \texttt{Obstruction} cards. 
These schemata elevate the live, trial-and-error process of proof exploration to first-class entities within the memory graph. 
By explicitly separating partial proof constructions (\texttt{PartialProof}) from falsified logical constraints (\texttt{Obstruction}), this layer preserves both valid intermediate subgoals and verified dead ends as structured, retrievable knowledge. 
Consequently, the architecture enables autonomous agents to navigate complex reasoning spaces and leverage accumulated exploratory experience without repeatedly rediscovering the same logical obstructions. 
The explicit taxonomy and data composition of all these cards are formalized in Table~\ref{tab:card_schemata}.

\input{app/Tikz_kb}
\begin{table}[t]
\small
\centering

\begin{subtable}{\linewidth}
\subcaption{Knowledge Base Cards}
\label{tab:card_schemata}
\begin{tabular*}{\linewidth}{@{\hspace{5pt}\extracolsep{\fill}} l l l @{\hspace{5pt}}}
\toprule
\textbf{Card Type} & \textbf{Core Function and Encoded Elements} & \textbf{Graph Connections} \\ 
\toprule
\texttt{Source} & External literature, including defs, thms, and proofs. & All cards related to the source \\
\texttt{Concept} & Cross-source math objects aggregating defs and formulas. & All cards including the concept \\
\texttt{Analysis} & Problem-solving insights, recurring proof patterns, etc. & \texttt{Concept}, \texttt{PartialProof}, \texttt{Obstruction} \\
\midrule
\texttt{Lean} & Formal artifacts tracking Lean declarations, proofs, etc. & \texttt{Concept}, \texttt{Source} \\
\texttt{PartialProof} & Incomplete proofs with partial structures and lemmas. & \texttt{Concept}, \texttt{Source}, \texttt{Obstruction} \\
\texttt{Obstruction} & Dead ends, structural anomalies, and logical conflicts. & \texttt{Concept}, \texttt{Source}, \texttt{PartialProof} \\
\bottomrule
\end{tabular*}
\end{subtable}

\begin{subtable}{\linewidth}
\subcaption{Specialist Roles in KB Manager}
\label{tab:kbmanager_role}
\begin{tabular*}{\linewidth}{@{\hspace{5pt}\extracolsep{\fill}} l l @{\hspace{5pt}}}
\toprule
\textbf{Role} & \textbf{Responsibility}\\ 
\toprule
Registrar & Registers sources into hash-addressed storage and tracks metadata before routing files to specialists.\\
Ingester & Compiles natural language sources into cards under a mandatory human confirmation gate. \\
Researcher & Queries the wiki graph to answer questions, persisting novel semantic syntheses.\\
Maintainer & Preserves graph health by repairing topological anomalies and adding necessary links.\\
Archivist & Integrates formal verification by mapping Lean 4 artifacts to cards and cross-indexing with concepts.\\
\bottomrule
\end{tabular*}
\end{subtable}

\caption{Knowledge Base Manager. (a) shows six kinds of Knowledge Base cards, their functions and connection roles. (b) shows five different subagents in the Knowledge Base Manager.}
\end{table}

Collectively, these cards form a heterogeneous \textit{mathematical dependency and evidence graph} governed by strict graph connection rules (detailed in Table~\ref{tab:card_schemata})~\citep{paulheim2016knowledge}.
Driven by this structural foundation, core system workflows are modeled directly as graph operations:
\textit{ingestion} initializes and links nodes, \textit{query} traverses bounded neighborhoods for contextual retrieval, \textit{maintenance} dynamically repairs broken or weak connections, and \textit{archiving} appends formal verification structures to the graph.
To guarantee the strict auditability required for safe mathematical reuse, the \textit{KB Manager} enforces a rigorous operational discipline that transparently tracks graph evolution while treating human modifications as the authoritative ground truth.
Ultimately, the \textit{KB Manager}'s contribution lies not merely in expanding context memory, but in establishing a mathematical card schema and a graph discipline that together render knowledge reusable, formalization-aware, and safe to compound across extended research sessions.
Refer to Figure~\ref{fig:kb} for an example of a knowledge base.

\subsubsection{Orchestrated Knowledge Lifecycle}
To materialize the harness architecture's core paradigms within the knowledge repository, the KB Manager replaces conventional linear extraction with a dynamic, state-isolated workflow. 
The system completely delegates specialized operations to a suite of functional experts while enforcing strict architectural boundaries to preserve global state inspectability.

Driven by our framework's emphasis on minimizing context overflow, this execution pipeline establishes rigid structural isolation by relying entirely on file-based handoffs rather than conversational history passing. 
Under this paradigm, each subagent is strictly restricted to a bounded operational scope to eliminate mutual cognitive interference (summarized in Table~\ref{tab:kbmanager_role}). 
Upon receiving raw materials, the \textit{Registrar} is invoked exclusively for physical intake—registering files into hash-addressed storage and tracking system metadata—thereby maintaining a clean separation between source indexing and downstream semantic processing. 
Guided by these generated manifests, the \textit{Orchestrator} dispatches control without transferring global memory: natural language prose is routed solely to the \textit{Ingester} for knowledge compilation, while formal scripts are isolated within the \textit{Archivist} for Lean 4 integration. 
For downstream knowledge retrieval, the \textit{Researcher}'s exploratory radius is similarly constrained, executing queries by strictly traversing the pre-compiled global index rather than conducting open-ended web searches. 
This strict compartmentalization of roles systematically prevents memory confusion among specialists throughout extended reasoning horizons.
Meanwhile, the \textit{Maintainer}'s independent topological audits guaranty global state inspectability throughout long-horizon reasoning.

Complementing these structural boundaries, the subagents' operational routines are further restricted by deterministic behavioral gates to eliminate the logical hallucinations and self-verification biases intrinsic to LLMs. 
Even within their isolated domains, agents are denied unilateral write privileges during graph-mutating operations. 
For instance, the \textit{Ingester} is prohibited from silently compiling parsed wiki cards; it is mandated to halt execution, summarize core takeaways, and await explicit human acknowledgment before final commit. 
Similarly, the \textit{Maintainer} and the \textit{Archivist} operate under a strict "propose-verify-commit" protocol. 
By transforming potential single-agent hallucinations into multi-stage verification obligations, these behavioral guardrails ensure that the continuously accumulated mathematical graph remains structurally sound and unconditionally trustworthy.

\section{Empirical Evaluation}
\label{sec:eval}
\begin{table}[t]
    \centering
    \small
    \setlength{\tabcolsep}{2pt}
    \begin{tabular*}{\linewidth}{@{\hspace{3pt}\extracolsep{\fill}} l l c c c @{\hspace{3pt}}}
    \toprule
    \textbf{Problems solved} & \textbf{Areas} & \textbf{Formal} & \textbf{Human} & \textbf{Ref.}\\
    \midrule
    Sparse Polynomial Divisibility Test is CoNP-hard & Algebraic Comp. Theory & \ding{55} & \ding{51} & \citep{qiu2026nphard}\\
    Output-sensitive Sparse Polynomial GCD is NP-hard & Algebraic Comp. Theory & \ding{117} & \ding{51} & \citep{cao2026conphard}\\
    Quasi-linear Time Multiplication of Sparse Polynomials & Algebraic Comp. Theory & \ding{117} & \ding{51} & \citep{huang2026quasilinear}\\
    Sparse Polynomials Exact-root Finding and Factorization & Algebraic Comp. Theory & \ding{55} & \ding{51} & \cite{HuangQL2026-ExactRoot}\\
    \cmidrule(lr){1-5}
    Katz's Problem on Generalized Airy Operators & Differential Algebra & \ding{55} & \ding{51} & \citep{cao2026equivalence}\\
    Skew Polynomial Multiplication & Differential Algebra & \ding{55} & \ding{51} & \cite{YeKe2026Skew}\\
    \cmidrule(lr){1-5}
    A GCD Criterion of Certain Binomial Coefficients & Number theory & \ding{51} & \ding{55} & \citep{guo2026gcd}\\
    An Integer is Sum of Tri, Pent, and Hept Number & Number theory & \ding{51} & \ding{55} & \citep{cao2026triangular}\\
    Removing non-CM Hypothesis for Shafarevich-Tate Bound & Number theory & \ding{55} & \ding{51} & -\\
    \cmidrule(lr){1-5}
    Ideal Equality for a Positive Non-Trace-Class Commutator & Operator Algebra & \ding{55} & \ding{55} & -\\
    \cmidrule(lr){1-5}
    Vasc's cyclic inequality for $n=9$ over positive reals & Inequality & \ding{55} & \ding{55} & \citep{guo2026finite}\\
    \bottomrule
    \end{tabular*}
    \caption{Summary of solved problems. To date, we have uploaded nine completed works to arXiv. The ``Human'' column indicates whether the work involved human participation, while the ``Formal'' column denotes whether it has been formally verified by FL-Prover. The symbol \ding{117} signifies partial formalization, where only the theorems presented in our paper are formalized, with all referenced theorems retained as \texttt{axiom}s.}
    \label{tab:results}
\end{table}

To evaluate the practical efficacy, mathematical adaptability, and closed-loop coordination capabilities of our framework, we deployed the system to address several long-standing open problems across Number Theory, Algebraic Complexity Theory, Differential Algebra, Operator Algebra, and Inequality, moving beyond static or standardized benchmarks. 
Table~\ref{tab:results} catalogs these solved problems by category. 
In addressing these open problems, \textit{MMAT} used strategic reasoning, symbolic computation, and formal verification across complex mathematical problems.
To comprehensively demonstrate these capabilities, the remainder of this section unfolds across three detailed case studies, each highlighting a distinct facet of our framework's performance.

\textbf{Full‑Cycle Theorem Proving Pipeline (\S\ref{sec:eval_gcd}).} 
This subsection details a comprehensive, end-to-end exposition of the entire systemic workflow using the open problem OEIS A287616 as a baseline case study. 
A proof emerges from the continuous interplay of natural language mathematical discovery, computational certificate validation, and iterative Lean 4 formalization, yielding a result that is rigorous, computationally verified, and formally certified~\citep{cao2026triangular}.

\textbf{Orchestrated Large-Scale Certification (\S\ref{sec:eval_vasc}).} 
Having established the system’s capacity to close the full proof cycle on a single problem, we now examine the robustness of its architecture when confronted with the combinatorial explosion that traditionally overwhelms automated provers.
The $n=9$ positive‑real case of Vasc’s cyclic inequality serves as a demanding testbed, requiring the system to navigate a vast, highly non‑uniform search space.
Here, the central Orchestrator robustly manages a primary verification scan in Python while simultaneously deploying parallel branches for speculative symbolic computation and fault‑tolerant exception handling. 
This design successfully validates all 40,320 distinct algebraic cones without a single failure~\citep{guo2026finite}.

\textbf{From a Single Problem to a Mathematical Project (\S\ref{sec:eval_prod}).} 
Moving beyond, we lastly examine the framework’s higher‑order capabilities to carry out a whole mathematical project, including proof-chain audit, theoretical reconstruction, and automated scientific discovery. 
For instance, in sparse polynomial multiplication, the system autonomously isolates and refutes a critical probability analysis flaw in the existing literature, with a counterexample at an astronomical scale ($N=10^{20}$).
Through a bidirectional human–AI co-reasoning loop, it then reconstructs a sound quasi-linear complexity framework~\citep{huang2026quasilinear}.
More importantly, it steers the broader project-level exploration toward more problems in sparse polynomials, including division, decomposition, GCD computation, factorization, and root finding.

\subsection{Full‑Cycle Theorem Proving Pipeline}
\label{sec:eval_gcd}
\begin{figure}
    \centering
    \includegraphics[trim=0 31 0 36,clip,width=.98\textwidth]{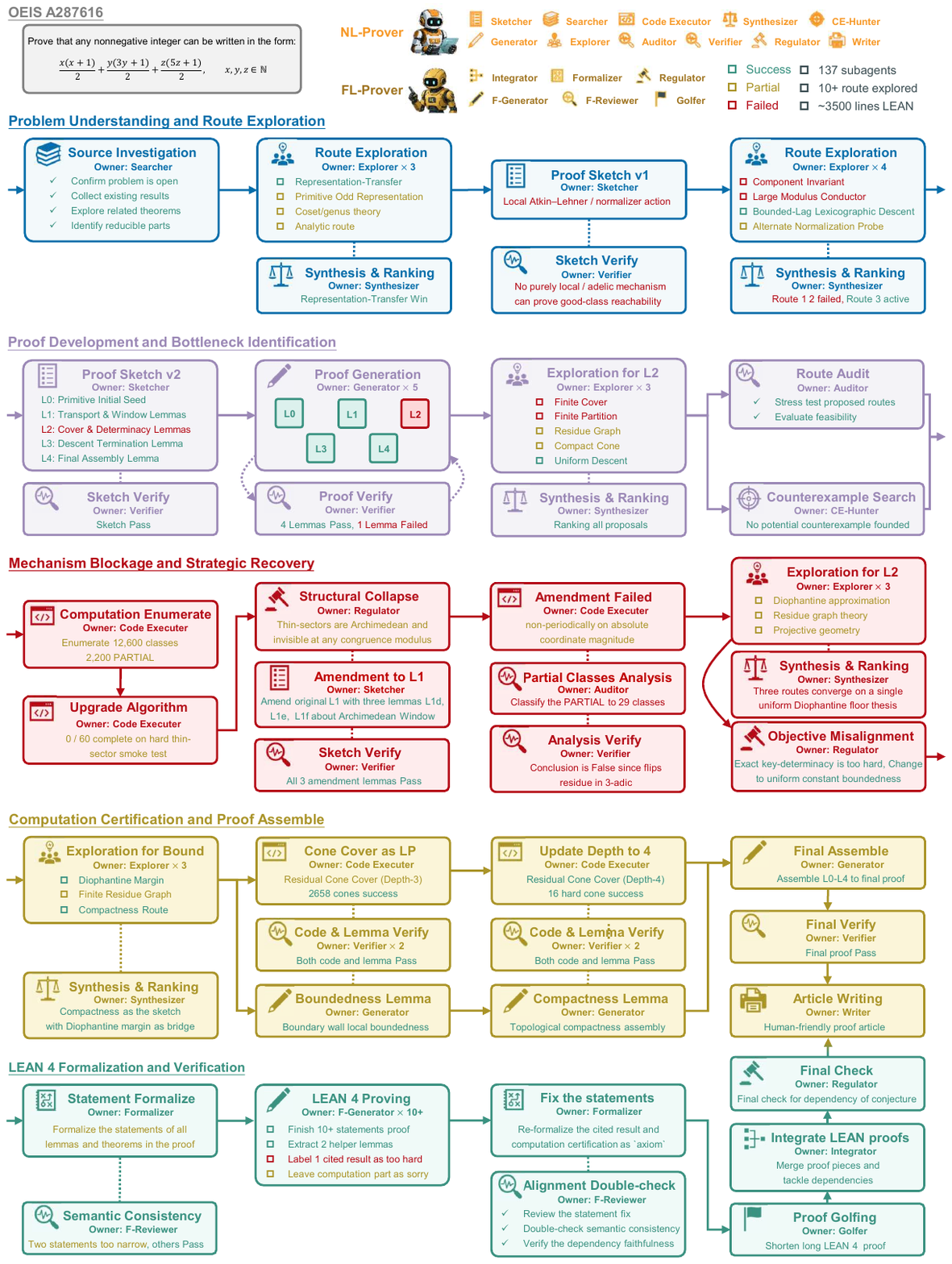}
    \caption{The multi-agent theorem proving pipeline example of \textit{MMAT} on open problem OEIS A287616. The NL-Prover orchestrates a series of 137 specialized subagents to execute route exploration, dynamic error recovery, large-scale computational validation, and finally finish the proof. FL-Prover compiles the proof into over 3,500 lines of formal Lean 4 code, providing rigorous guarantees for the natural-language proof.}
    \label{fig:workflow}
\end{figure}

We use the open problem OEIS A287616 as a case study to demonstrate the complete theorem-proving pipeline of the \textit{MMAT}. The problem asks whether every nonnegative integer $n$ can be represented in the form
\begin{equation}
    n=\frac{x(x+1)}{2}+\frac{y(3y+1)}{2}+\frac{z(5z+1)}{2},
\qquad x,y,z\in\mathbb N.
\end{equation}



To solve this problem, the NL-Prover constructs the whole proof by invoking 137 subagents (Figure~\ref{fig:workflow}). Starting from classical quadratic-form theory, the system progressively develops the Primitive Seed, Move Calculus, Elementary Descents, Residual Gap, and subsequent reduction lemmas with multiple specialized subagents, including the \textit{Searcher}, \textit{Explorer}, \textit{Sketcher}, \textit{Generator} and \textit{Verifier}. Once the proof is reduced to the Residual Cone, multiple mathematical directions—including Finite Partition, Finite Residue Graph, Compact Projective Cone, Finite Cover, and Uniform Descent—are explored in parallel. The \textit{Code Executor} automatically converts these mathematical hypotheses into executable programs, leveraging its symbolic component in Python constructing Cover Certificates and Exact-LP Certificates through large-scale computation, while the \textit{CE-Hunter} continuously searches for adversarial configurations that may invalidate the current theory.

The flexibility of the framework becomes most evident during the exploration process. Rather than following a single proof strategy, the system repeatedly recovers from failed mathematical routes. For example, the initial objective is to establish \emph{key determinacy}, namely that a finite state key uniquely determines the descent path. However, both mathematical analysis and computational evidence show that this objective is false. The \textit{Regulator} therefore reformulates the mathematical target as \emph{boundedness}, requiring only that every residual state descends within a uniformly bounded number of steps, without demanding a unique finite-state description of the descent. The proof subsequently undergoes multiple recovery stages, including quantifier-order correction, cover-certificate reconstruction, modulus refinement, proof-depth extension ($\ell_W:4\rightarrow5$), and Exact-LP certification, ultimately converging to the complete natural-language proof.

After the natural-language proof is completed, the FL-Prover automatically formalizes the proof in over 3,500 lines of Lean 4 code. Rather than serving as a one-way translation procedure, formalization itself forms another iterative verification loop. The \textit{Formalizer} first generates Lean theorem skeletons, after which the \textit{F-Reviewer} checks their consistency against the mathematical manuscript. During the first review, the computational verification is found to be encapsulated by an overly broad axiom boundary. The task is therefore returned to the \textit{Formalizer}, who progressively refines the formalization boundary until only two explicit external assumptions remain: the classical genus local-global theorem and the computer-certified finite-cover statement \texttt{fin\_cell\_residual}. All remaining mathematical components—including the Primitive Seed, Move Calculus, Residual Gap, Reduction, and Final Assembly—are fully formalized in Lean4. Subsequently, the \textit{F-Generator}, \textit{Integrator}, \textit{Golfer}, and \textit{Regulator} complete theorem proving, repository integration, code refinement, and global dependency auditing. Finally, the system invokes \texttt{\#print axioms} to confirm that no unintended project axioms remain.


This case study demonstrates \textit{MMAT} as an end-to-end pipeline spanning from natural language to formal code. By unifying mathematical discovery, computational validation, and formal verification (FL-Prover), the system creates a closed, self-reinforcing research loop. This synergy ensures that complex proofs successfully converge into results that are simultaneously mathematically rigorous, computationally verified, and formally certified.

\subsection{Orchestrated Large-Scale Certification}
\label{sec:eval_vasc}

We evaluate our framework on the $n=9$ positive-real case of Vasc's cyclic inequality:
\begin{equation}
    C_{9}(x)=\sum_{i=1}^{9}\frac{x_{i}-x_{i+1}}{x_{i+1}+x_{i+2}} \ge 0.
\end{equation}
The main difficulty is that no single sum-of-squares or Arithmetic Mean-Geometric Mean (AM-GM) identity certifies nonnegativity over the whole positive orthant.
The domain must be partitioned into finitely many regions, each with its own certificate. 
Through automated theorem retrieval, \textit{MMAT} first surveyed existing methods and found that Successive Difference Substitution (SDS) successfully handles $n\leq7$~\citep{xia2016automated,sil2023mse}. 
However, for $n=9$ the $8!=40{,}320$ cones make a direct reproduction of the lower-order pattern impossible without an automated, adaptive search.

To overcome this, a sketcher-generator-verifier workflow (Figure \ref{fig:vasc_cert}) produces a trustworthy proof plan that splits the argument into algebraic reductions and a case-by-case certification.  
The algebraic reduction clears the positive denominators, yielding a homogeneous polynomial $P_9$.
Then it uses cyclic symmetry to place a maximal coordinate first, followed by sorting and reparametrization via cumulative gaps (SDS substitution) to map each cone onto a nonnegative orthant.  
The case certification computes the polynomial coefficients for every reduced cone: if all coefficients are strictly positive, the cone is immediately certified; otherwise, a Polya-multiplier method is applied to obtain a nonnegativity witness.  
A code executor then carries out this plan with an automated Python pipeline, verifying the $40{,}320$ cones in batches of $64$ cases.

\input{app/Tikz_vasc}

During the main brute-force computation, some patterns emerge when 1,152 samples are verified. 
Thus, the \textit{Orchestrator} opens a speculative branch that searches for symbolic invariants able to simplify the verification, as shown in the green box of Figure~\ref{fig:vasc_branch}. 
Within this branch, \textit{Explorers} scan verified data snapshots to propose pattern simplifications, which the \textit{Sketcher} formulates into structured algebraic hypotheses. 
Rather than adopting these candidate rules directly, the \textit{Orchestrator} routes them to the \textit{Generator} for proof attempts and to the \textit{CE-Hunter} for falsification tests. 
Unfortunately, a structural obstruction falsifies the generalized hypothesis, so the \textit{Orchestrator} abandons the global symbolic-expansion route. 
Because this branch runs in parallel, these explorations and the eventual rollback do not interrupt the primary verification scan.

Conversely, when the certificate checker leaves a case unresolved (sample 5,916), the \textit{Orchestrator} pauses the primary scan and opens a fault-tolerant recovery branch, as illustrated in the red box of Figure~\ref{fig:vasc_branch}.
This mitigation branch first dispatches the \textit{CE-Hunter} to probe the unresolved case for violations. 
Finding no counterexamples does not establish nonnegativity, but it signals that the case warrants a dedicated certificate, so the \textit{Orchestrator} resumes the primary verification pipeline on the remaining cases.
While the primary scan continues, a \textit{Generator} synthesizes a proof patch for the skipped case. 
This patch is then routed through the \textit{Verifier} to authenticate its mathematical soundness, and subsequently through the \textit{Code Executor} to check by code.
After regenerating certificate for the case, the exception is successfully resolved.

Through this orchestrated process, the complete fixed-maximum certificate over the entire lexicographic range $[0,40320)$ passed both the mechanical checker and an independent computational audit.
A subsequent refinement pass produced a shorter proof that was re-verified and adopted, yielding a fully certified proof of the positive-real $n=9$ Vasc inequality.  
The final certificate classifies every terminal row into one of three leaf types: 36,815 positive-coefficient leaves, 2,236 Polya-multiplier leaves, and 1,269 leaves certified by AM-GM–based Sum of Squares (SOS).
Refer to \citep{guo2026finite} for the complete proof.

\subsection{From a Single Problem to a Mathematical Project}
\label{sec:eval_prod}
Having confirmed the framework's capability in full-cycle theorem proving and robustness in large-scale machine certification, we now shift the focus to intellectual depth, demonstrating how \textit{MMAT} can drive an entire research project. 
Sparse polynomial computation serves as a running case study in this section.

Sparse polynomials are the standard representation in practice and underpin computer algebra systems such as Maple and Mathematica, which are widely used in scientific and engineering computation.
Surprisingly, for many basic problems in sparse polynomials, including multiplication, division, decomposition, GCD computation, factorization, and root finding, their computational complexity and optimal algorithms remain poorly understood~\citep{plaisted1984,SparsityChallenges,Roche2018}.

We begin with the first open problem in \citep{Roche2018}, which asks for a quasi-linear sparse polynomial multiplication algorithm. 
A positive solution is given over integers in~\citet{giorgi2020essentially}, but an automatic audit by \textit{MMAT} uncovered a critical logical flaw in their probabilistic analysis.
Subsequently, \textit{MMAT} constructed a counterexample, conclusively invalidating the claim.
In response, a completely new proof was constructed through iterative human–AI collaboration, and its correctness was ultimately machine‑checked by Lean 4~\citep{huang2026quasilinear}.
With this project underway, we further examined more problems for sparse polynomials and resolved four long-standing open problems (Figure \ref{fig:sparse_poly})~\citep{huang2026quasilinear,cao2026conphard,qiu2026nphard,HuangQL2026-ExactRoot}, among which \textit{MMAT} provides a rigorous proof of the key theorem.
The remainder of this section follows this progression, from automated error detection to systematic exploration of the topic.

\subsubsection{Proof-chain Auditing}
The discovery of the logic flaw in Lemma 4.7(ii) represents a paradigm shift: an interactive multi-agent system can systematically audit intricate proof dependencies and aggressively navigate extreme parameter spaces to falsify seemingly ironclad theoretical claims.

In theoretical mathematics, primary conclusions rest on a dense, hierarchical web of supporting lemmas imported from external literature, which are frequently accepted as unverified ``black boxes''. 
\textit{MMAT} addresses this through automated proof-chain auditing.
As illustrated in the sequential pipeline of Figure~\ref{fig:proofchain}, the \textit{Orchestrator} parallelized the team's execution. 
The \textit{Searcher} (SEAR) traced upstream dependencies to audit prerequisite assumptions within Huang's UIPOLY framework~\citep{huang2019sparse}. 
Simultaneously, the \textit{Auditor} (AUD) dissected micro-algebraic boundaries, including the intermediate support size $T_p$ and the ordinary product $F_pG_p$.
This synchronized deep-dive allowed the agents to successfully isolate a critical logical flaw and trigger an automated \texttt{HALT} sequence precisely at Lemma~4.7(ii).

To initiate a concrete refutation, the \textit{Orchestrator} pivoted operations toward counterexample exploration, empowering the \textit{CE-Hunter} to spearhead the search under a hypothesis testing framework formalized by the \textit{Sketcher}. 
Through this setup, the \textit{CE-Hunter} identified that massive scales cause canceled terms in the ordinary product $F_pG_p$ to split violently upon modulo reductions, triggering an illegal expansion of the intermediate support set $T_p$.
Leveraging this insight, the \textit{CE-Hunter} systematically swept the parameter space and locked onto a highly unexpected configuration at $N=10^{20}$, as displayed in the counterexample block in Figure~\ref{fig:sparse_poly_ce}.
Such extreme scales lie beyond the reach of conventional verification, enabling the discovery of a structural mutation.

To ground this symbolic insight into definitive proof, the \textit{Code Executor} performed a parallel Python‑based finite enumeration.
As detailed in the green block in Figure~\ref{fig:sparse_poly_ce}, the actual success probability under this extreme scale plummets to $\mathbb{P} = \sfrac{7495}{30182} \approx 0.248$, directly refuting the original paper's theoretical claim of $1-\sfrac{\mu_1}{2} = 3/4$. 
Backed by this data, the \textit{Verifier} authenticated the algebraic model, the \textit{Regulator} issued the dual-consensus verdict, and the \textit{Writer} seamlessly compiled the formal refutation report.

\input{app/Tikz_counterexample}
\subsubsection{Human-MMAT Interaction Proof}
With the logic flaw isolated and refuted, the pipeline shifted to theoretical reconstruction, establishing a framework for quasi‑linear bit complexity over integers through interactive human–AI collaboration. 
This collaboration formed a bidirectional, self‑correcting loop, as the following examples illustrate.

At the finest level, the agent initially conflated the absorption rules of soft‑O notation, incorrectly reducing \(m\log n\) to \(\widetilde{O}(m)\) by false analogy with the valid \(n\log n = \widetilde{O}(n)\). 
Once the human mathematician identified the error, the agent immediately corrected the derivation and stored the corrected rule, preventing the same conflation from recurring in subsequent proofs.

At a higher strategic level, the human mathematician proposed deducing the sparse multiplication theorem over finite fields \(\mathbb{F}_q[x]\) from the integer case \(\mathbb{Z}[x]\) via reduction. 
The agent successfully proved the required lemmas, verified their compatibility, and confirmed the approach was viable.
The resulting validated structures were then ingested by the \textit{KB‑Manager} as reusable building blocks. 
Had this direction proved infeasible, the agent would have automatically assumed a skeptical role, constructing explicit algebraic counterexamples to disprove the hypothesis. 
This negative result would have been relayed to the human to realign the research direction, and the \textit{KB‑Manager} would have simultaneously archived the counterexample as an experiential constraint, preventing the collective intelligence from revisiting the same blind alley.

\subsubsection{Project-level Exploration}
The resolution of the sparse multiplication problem served as an epistemic catalyst for a systematic investigation across the landscape of sparse polynomial computation, with consolidated knowledge structured by the KN Manager.
Figure~\ref{fig:kb} presents an example of how knowledge about sparse polynomial computation, as organized by the KB manager \textit{MMAT}, was progressively mapped over the course of this project.
Figure~\ref{fig:sparse_poly} illustrates the six interrelated problems that were solved. We briefly summarize each problem and the corresponding contribution in the following.
\input{app/Tikz_progress}

\begin{itemize}[leftmargin=*,nosep]
\item \textbf{Multiplication.} 
Roche's Open Problem 1 on quasi‑linear complexity of multiplication~\cite{Roche2018} was settled over the integers~\citep{huang2026quasilinear}.
\item \textbf{Divisibility.} 
The longstanding open problem of the complexity of sparse polynomial divisibility testing~\cite{plaisted1984,SparsityChallenges,Roche2018} was solved over finite fields by showing that it is $\text{coNP}$‑hard under BPP~\cite{cao2026conphard}.
\item  \textbf{GCD.} 
The longstanding problem of output‑sensitive GCD computation posed by Davenport et al.~\cite{SparsityChallenges} has been resolved over finite fields and demonstrated to be NP‑hard under BPP reductions~\citep{qiu2026nphard}.
    %
    \item 
    \textbf{Decomposition.} 
    Extending the classical conjecture of \citet{Erdos1949} and its univariate resolutions by \citet{Schinzel1987} and \citet{Zannier2008},  the multivariate version is solved: sparsity of $g^e$ implies sparsity of $g$ and  a polynomial‑time algorithm  for $g$ is given~\citep{HuangQL2026-ExactRoot}. 

    \item \textbf{Roots.} 
 Extending the known NP‑hardness of detecting a root in a finite field~\cite{BI-SIAMJC2016}, it was established that deciding whether a sparse polynomial vanishes at a root of unity is NP‑hard, further tightening the complexity landscape~\citep{qiu2026nphard}.

    \item \textbf{Factorization.} 
    It is proved that sparsity of $gh$ leads to the sparsity of $g$ and $h$, provided $h$ is multilinear~\citep{HuangQL2026-ExactRoot}, advancing the line of inquiries by \citet{GathenKaltofen1985} and \citet{BSV2020}.
\end{itemize}

Thus, \textit{MMAT} moves decisively beyond solving isolated problems: it acts as an intellectual architect, steering a single problem toward a comprehensive project that encompasses an entire mathematical domain.

\section{Conclusion}

In this paper, we introduced the \textit{MechMath Agent Team (MMAT)}, a multi‑agent system designed to serve as a co‑pilot across the full cycle of mathematical research. 
Central to MMAT is the tripartite Harness Architecture, which decouples system responsibilities into \textit{Control}, \textit{Execution}, and \textit{Augmentation planes}, thereby reconciling rigorous logical control with the exploratory agility demanded by open‑ended problems. 
Three specialized agents-a \textit{Knowledge Base Manager}, a \textit{Natural Language Prover}, and a \textit{Formal Language Prover} leveraging the Lean 4 compiler—operate in a closed loop to produce formally certified proofs. 
We evaluated \textit{MMAT} on open problems spanning number theory, algebraic complexity theory, differential algebra, and inequalities; over a two‑month deployment, it solved ten problems, including Katz’s question on generalized Airy operators, conjectures on sparse polynomial divisibility and multiplication, and a multivariate extension of the Erdős–Rényi conjecture. 
These results confirm that \textit{MMAT} can audit existing proofs, incorporate human feedback, and steer exploration toward uncharted mathematical directions. 
This work demonstrates that structured multi‑agent collaboration, when combined with formal verification and human–AI co‑reasoning, can meaningfully accelerate mathematical discovery, opening new avenues for AI‑assisted research.

\section*{Acknowledgment}
This paper is supported by the Strategic Priority Research Program of CAS Grants XDA0480502 and XDA0480503, and NSFC Grant No. 12288201.

\bibliographystyle{KLMM/klmm}   
\bibliography{refs} 

\end{document}